\documentclass[journal]{IEEEtran}
\usepackage{times}
\usepackage{epsfig}
\usepackage{graphicx}
\usepackage{amsmath}
\usepackage{bm}
\usepackage{subeqnarray}
\usepackage{cases}
\usepackage{amssymb}
\newcounter{defcounter}
\usepackage{pdfpages}
\usepackage{color}
\usepackage{multirow}
\usepackage{booktabs}
\usepackage{array}
\usepackage[normalem]{ulem}
\usepackage{array,multirow}

\usepackage{latexsym}
\usepackage[utf8x]{inputenc}
\usepackage{amssymb,amsmath,amsfonts}
\usepackage{algorithm}
\usepackage{algpseudocode}
\usepackage{lineno}
\usepackage{color}
\usepackage{url}
\usepackage{amsthm}

\newcommand{\cut}[1]{}

\begin{document}
%
\title{Latent Constrained  Correlation Filter}
%
%
%

\author{Baochang~Zhang,
Shangzhen~Luan,
Chen~Chen,
Jungong~Han,
Wei~Wang,
Alessandro~Perina,
and~Ling~Shao,~Senior~Member~IEEE
\thanks{The work was supported in part by the Natural Science Foundation of China under Contract 61672079 and 61473086.  The work of B. Zhang was supported in part by the Program for New Century Excellent Talents University within the Ministry of Education, China. (B. Zhang and S. Luan contributed equally to this work.)~   (Corresponding author: Jungong Han.) }
\thanks{B. Zhang, S. Luan and W. Wang are with the School of Automation Science and Electrical Engineering, Beihang University, Beijing, China. Email: bczhang@buaa.edu.cn.}
\thanks{C. Chen is with Center for Research in Computer Vision (CRCV), University of Central Florida, Orlando, FL, USA. Email: chenchen870713@gmail.com.}
\thanks{J. Han is with the School of Computing and Communications, Lancaster University, Lancaster LA1 4YW, U.K. Email: jungonghan77@gmail.com.}
\thanks{A. Perina is with  Microsoft Corporation, Redmond, WA, USA.}
\thanks{L. Shao is with the School of Computing Sciences, University of East Anglia, Norwich NR4 7TJ, U.K. Email: ling.shao@ieee.org.}
\thanks{Manuscript received XX XX, 2016.}
}

%
%

\markboth{IEEE Transactions on Image Processing}%
{Shell \MakeLowercase{\textit{et al.}}: Bare Demo of IEEEtran.cls for IEEE Journals}
%



\maketitle

\begin{abstract}
Correlation filters are special classifiers designed for shift-invariant object recognition, which are robust to pattern distortions. The recent literature shows that combining a set of sub-filters trained based on a single or a small group of images obtains the best performance. The idea is equivalent to estimating variable distribution based on the data sampling (bagging), which can be interpreted as finding solutions (variable distribution approximation) directly from sampled data space. However, this methodology fails to account for the variations existed in the data. In this paper, we introduce an intermediate step -- solution sampling -- after the data sampling step to form a subspace, in which an optimal solution can be estimated.  More specifically, we propose a new method, named latent constrained correlation filters (LCCF), by mapping the correlation filters to a given latent subspace, and develop a new learning framework in the latent subspace that embeds distribution-related constraints into the original problem. To solve the optimization problem, we introduce a subspace based alternating direction method of multipliers (SADMM), which is proven to converge at the saddle point. Our approach is successfully applied to three different tasks, including eye localization, car detection and object tracking. Extensive experiments demonstrate that LCCF outperforms the state-of-the-art methods. \footnote{The source code will be publicly available. https://github.com/bczhangbczhang/}
\end{abstract}


	%
	%
    \maketitle		

	\IEEEdisplaynotcompsoctitleabstractindextext


	%
	\IEEEpeerreviewmaketitle

\section{Introduction}
\label{sec:introduction}
	
Correlation filters have attracted increasing attention due to its simplicity and high efficiency. They are usually trained in the frequency domain with the aim of producing a strong correlation peak on the pattern of interest while suppressing the response to the background. To this end,  a regression process is usually used to obtain a Gaussian output that is robust to shifting. Recently, correlation filters have emerged as a useful tool for a variety of tasks such as object detection and object tracking.

The correlation filter method is first proposed by Hester and Casasent, named synthetic discriminant functions (SDF) \cite{sdf}, which focuses more on formulating the theory. Later on, to facilitate more practical applications, many variations are proposed to solve object detection and tracking problems.
For object detection task, early research can be traced back to \cite{mace}, where Abhijit \textit{et al.} synthesize filters by Minimizing the Average Correlation plane Energy (MACE), thus allowing easy detection in the full correlation plane as well as control of the correlation peak value.  In their improved work \cite{mach}, it was noted that the hard constraints of MACE cause issues with distrotion tolerance. Therefore, they eliminate the hard constraints and require the filter to produce a high average correlation response instead. A newer type of CF, named Average of Synthetic Exact Filters (ASEF) \cite{asef}, tunes filters for particular tasks, where ASEF specifies the entire correlation output for each  training image, rather than specifying a single peak vaule used in earlier methods. Despite its better capability of dealing with the over-fitting problem, the need of a large number of training images makes it difficult for real-time applications. Alternatively, Multi-Channel Correlation Filters (MCCF) \cite{mccf} take advantage of multi-channel features, such as Histogram of Oriented Gradients (HOG) \cite{hog} and Scale-Invariant Feature Transform (SIFT) \cite{sift}, in which each feature responds differently and the outputs are combined to achieve high performance. For the tracking task, Minimum Output Sum of Squared Error filters (MOSSE) \cite{mosse} are considered as the earliest CF tracker, which intends to produce ASEF-like filters from fewer training images. Its essential part is a mapping from the training inputs to the desired training outputs by minimizing the sum of squared error between the actual output of the convolution and the desired output of the convolution. Alternatively, Kernelized Correlation Filters (KCF) \cite{kcf} map the feature to a kernel space, and utilize the properties of the cyclic matrix to optimize the solution process. Since then, most trackers are improved based on KCF, such as \cite{liu2015part,kpcf,multikernel,longterm,oct}. We provide a comprehensive review of these approaches in the next section.

In general, the existing correlation filtering algorithms work pretty well in ideal situations. However, the performance degrades dramatically when dealing with distorted data, such as occlusion, noise, illumination, and shifting. Adding constrains is a sensible way to improve robustness of classification, leading to the constrained correlation filters.  In \cite{boundary}, a new correlation filter considering the boundary effect constraint can greatly reduce the number of examples involved in a correlation filter that are affected by boundary effects. The maximum margin correlation filters (MMCF) \cite{ccf1}, constraining the output at the target location, show better robustness to outliers. The Distance Classifier Correlation Filters (DCCF) \cite{ccf2} incorporate the distance information into the filter calculation for multi-class tasks. In our previous work \cite{pcm}, an adaptive multi-class correlation filters (AMCF) method is introduced based on an alternating direction method of multipliers (ADMM) framework by considering the multiple-class output information in the optimization objective.
\begin{figure*}
	\includegraphics[width=0.85\textwidth]{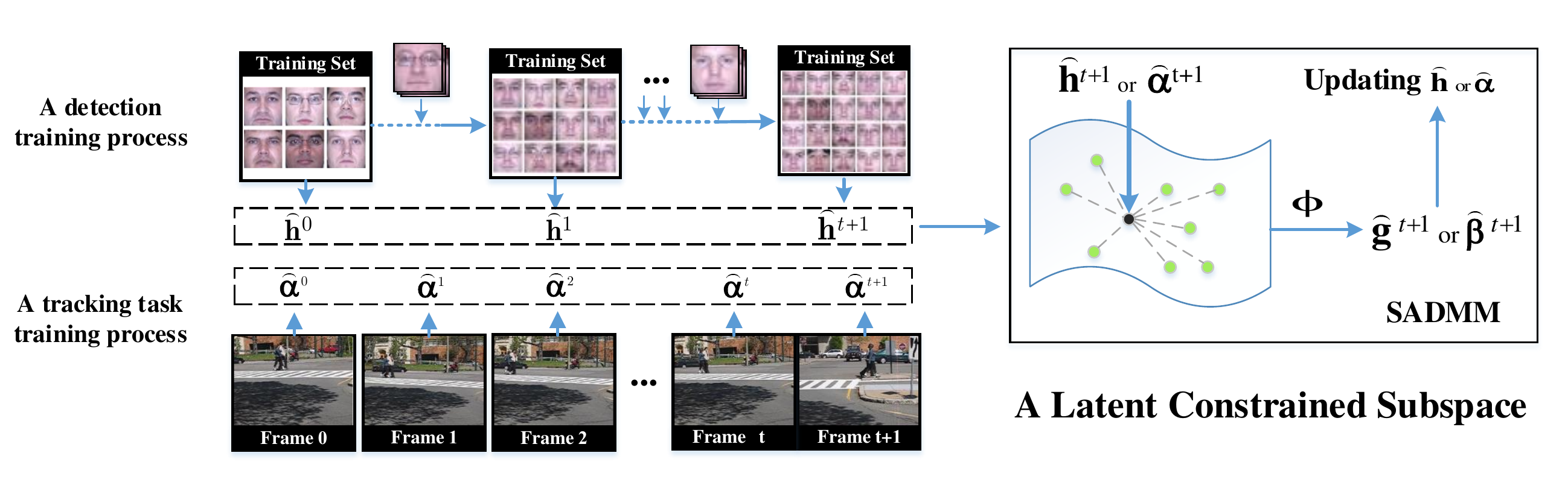}
	\vspace{-0.1cm}
	\caption{The framework of the proposed latent constrained correlation filters (LCCF). In the left-upper part, we show a detection training process. $\hat{\mathbf{h}}^{0:t+1}$ forms a subspace, in which $\hat{\mathbf{g}}^{t+1}$ is obtained based on a projection $\Phi$. In the left-bottom part, we show a tracking process. $\hat{\bm{\alpha}}^{0:t+1}$ forms a subspace, in which  $\hat{\bm{\beta}}^{t+1}$ is obtained based on a projection $\Phi$. This procedure leads to a Subspace based Alternating Direction Method Of Multipliers (SADMM) algorithm.}
	\vspace{-0.2cm}
	\label{fig:figframe}
\end{figure*}	
	
\textbf{Problem}: Given the training samples, at the core of correlation filtering is to find the optimal filters, which requires the unknown variable distribution estimation. Traditional algorithms normally adopt one of the following schemes: 1) finding a single filter (i.e., channel filter \cite{mccf}) trained by a regression process based on all training samples \cite{mccf,boundary}, and 2) finding a set of sub-filters (a single filter per image \cite{asef}) and eventually integrating them into one filter. Here, the combination can be either based on averaging the sub-filters in an off-line manner \cite{asef} or an on-line iterative updating procedure \cite{kcf}, which is similar to that of the bagging method \cite{bagging}. Revealed by the literature, the performance of the second scheme is better than that of the first one \cite{kcf}, even though it is  computationally more expensive. The second scheme is equivalent to estimating solution distribution based on only a limited amount of sampled data, which fails to consider the variations existed in the data in the optimization objective.
	
One fact that has been overlooked in correlation filter learning is that data sampling (bagging) can actually lead to solution sampling, which is traditionally used to find an ensemble classifier. However, we argue that the bagging results can also be used to estimate the distribution of the solutions. Then, the distribution (later subspace) in turn can be deployed to constrain and improve the original solution. In this paper, we attempt to implement the above idea in correlation filtering, in order to enhance the robustness of the algorithm.

The framework of the proposed Latent Constrained Correlation Filters (LCCF) is shown in Fig. 1. To find the solution sampling in the training process, unlike an ad-hoc way that directly inputs all samples to train correlation filters, we train sub-filters step by step based on iteratively selecting subsets. Instead of estimating a real distribution for an unsolved variable that is generally very difficult, we exploit sampling solutions to form a subspace, in which the bad solution from a noisy sample set can theoretically be recovered after being projected onto this subspace in an Alternating Direction Method of Multipliers (ADMM) scheme. Eventually, we can find a better result from the subspace (subset) that contains different kinds of solutions to our problem. From the optimization perspective, the subspace is actually used to constrain the solution, as shown in Fig. 1. In fact, the above constrained learning method is a balanced learning model across different training sets. The application of constraints derived from data structure in the learning model is capable of achieving good solutions \cite{chang2007guiding,cabanes2010learning,mct}. This is also confirmed by \cite{cabanes2010learning}, in which the topological constraints are learned by using data structure information. In \cite{mct}, Zhang \textit{et al.} put forward a new ADMM method, which can include manifold constraints during the optimization process of sparse representation classification (SRC). These methods all achieve promising results by adding constraints.
	
Another key issue is how to efficiently embed the subspace constraints in the optimization process. In this paper, we propose a Subspace based Alternating Direction Method of Multipliers (SADMM). The classical ADMM is an algorithm that solves convex optimization problems by breaking them into smaller pieces, and each of which can be handled easily \cite{admm}. However, the original ADMM cannot be directly applied to solve our problem due to its infeasibility of handling the subspace constraint. In contrast,  the proposed SADMM is more flexible and proved to converge at the saddle point, therefore enabling a faster algorithm. In summary, our LCCF based on SADMM differ from the previous approaches in two aspects:

1) Our SADMM algorithm takes advantage of the inherent visual data structure for solving the optimization problem. We show that SADMM theoretically converges at the saddle point in an efficient way.

2) Our SADMM can be used to solve both linear and kernerlized correlation filtering based on a latent subspace constraint. Experimental results show that it consistently outperforms the state-of-the-art methods on three different tasks, i.e., eye localization, car detection and object tracking, revealing the generalization ability of the proposed SADMM model.


\textbf{Notation}: In this paper, scalars are represented by Italic letters (e.g., $B$), vectors are presented in lowercase boldface (e.g., $\mathbf{x}$), and matrices and supervectors are in upper case boldface (e.g., $\mathbf{X}$). Sign $\hat{}$ represents the Fourier form of a variable (e.g., $\hat{\mathbf{h}}$ is the Fourier form of $\mathbf{h}$). $T$ is the transpose operator of matrix. The operator $\textit{diag}$ converts the $D$ dimensional vector into a $D \times D$ dimensional matrix, which is diagonal with the original vector. The subscript $i$ represents the $i^{th}$ element in a data set (i.e., $\mathbf{x}_i$ refers to the $i^{th}$ sample in a training set  or a test set), the subscript $[k]$ represents channel (i.e., $\mathbf{x}_{i,[k]}$ represents the $k^{th}$ channel of $\mathbf{x}_{i}$, ${\bf h}_{[k]}$ refers to the $k^{th}$ channel of $\bf h$ ), and the superscript refers to the iterations of variable (i.e., $\hat{\bf h}^{t}$ for the variable $\hat{\bf h}$ in the $t^{th}$ iteration). For clarity, we summarize main variables in Table \ref{tab:variables}.
\begin{table*}
	\caption{A brief description of variables and operators used in the paper.}
	\centering
	\begin{tabular}{|ll|}
		\hline
		$\bf h$: correlation filter for linear case  & $\bf g $: the mapping of $\bf h$ in a latent subspace \\
		$\bm \alpha$: correlation filter for the kernelized case  &  $\bm \beta$:  the mapping of $\bm \alpha$ in a latent subspace\\
		$[k]$: channel index & $K $: the number of channels    \\
		$t $: iterative index &  $T $: the number of previous tracking frames used in subspace \\
		
		$B $: the number of samples in the training set
		& $maxiter$: the number of subsets for the training set	     \\
		\hline	
		
		$E({\bf h})$: a general objective function of {\bf h}  &    ${E}({\bm\alpha})$: the objective function for KCF based on SADMM\\
		$E_{S,L}({\bf{h}})$: the objective function for linear correlation filter based on SADMM &
		\\
	    $E_S({\bf{h}}) $: the objective function obtained based on a subspace contraint and $E({\bf h})$&\\
		\hline
	\end{tabular}
	\label{tab:variables}
\end{table*}

\section{related work}
Comparing with the traditional object detection and tracking algorithms \cite{hog,tld,struck,featureselection,zhang2015robust}, correlation filtering exploits convolution to simplify the mapping between the input training image and the output correlation plane, and has high computational efficiency and strong robustness. A flurry of recent extensions to correlation filter have been successfully used in the object detection and tracking applications.
	
Bolme \textit{et al.} propose a method to learn several accurate weak classifiers to construct a strong classifier for eye localization \cite{chang2007guiding}. Later, they propose to learn a Minimum Output Sum of Squared Error (MOSSE) filter \cite{cabanes2010learning} for visual tracking on gray-scale images, which is very efficient with a speed reaching several hundreds frames per second. Heriques \textit{et al.} present a method based on kernel ridge regression, exploiting a  dense sampling strategy and the circulant structure to simplify the training and testing processes \cite{csk}. The kernel ridge regression based method has been further improved in recent years and many variants appear \cite{kcf,kpcf,multikernel,color,longterm}. By using HOG features, KCF is developed to improve the performance of CSK \cite{kcf}. Tang \textit{et al.} \cite{multikernel} introduce the multi-kernel correlation filter which is able to take advantage of the invariance discriminative power spectrums of various features. Part-based correlation filters, such as \cite{liu2015part,yao2016exploiting} adopt the correlation filters as part classifiers to effectively handle partial occlusions. Danelljan \textit{et al.} exploit the color attributes of a target object and learn an adaptive correlation filter by mapping multi-channel features into a Gaussian kernel space \cite{color}. In \cite{dsst}, Discriminative Scale Space Tracker (DSST) is proposed to handle scale variations based on a scale pyramid representation. This scale assessment method can also be embedded into other models \cite{multikernel,longterm}. Ma \textit{et al.} develop a re-detecting process to further improve the performance of KCF \cite{longterm}. In \cite{boundary}, Galoogahi \textit{et.al} present a method to limit the circular boundary effects while preserving many of the computational advantages of canonical frequency domain correlation filters. In \cite{cfcf}, convolutional features correlation filters (CFCF) exploit features extracted from deep convolutional neural networks (DCNN) trained on object recognition datasets to improve tracking accuracy and robustness.
	
From the review of previous works, it can be seen that the  distribution introduced by the bagging method is not well studied for the correlation filter calculation. However, the distribution information is important to calculate robust filters, especially when the data suffer from severe noise, occlusion, etc.
%
	%
		
\section{Subspace based alternating direction method of multipliers (SADMM)}
\label{sec:proposed}
		
The {Augmented Lagrangian Multiplier (ALM) methods are a  class of algorithms for solving constrained optimization problems by including penalty terms to the objective function \cite{alm}. As a variant of the standard ALM method that uses partial updates (similar to the Gauss-Seidel method for solving linear equations), ADMM recently gained much more attention due to its adaptability to several problems \cite{admm}. By solving iteratively a set of simpler convex optimization sub-problems, each of which can be handled easily.} In this section, we show how the visual data structure, i.e., subspace constraints, can be embedded into the ALM minimization to define SADMM. We then present the resulting algorithm for the proposed SADMM and the solution of each sub-problem.
			
The primary task is to minimize $	E(\hat{\bf h})$ that is a general and convex optimization objective. In order to exploit the property of solution sampling from the data sampling, the subspace constraint is added to the original optimization problem. That is to say, instead of estimating a real distribution function of any unsolved variable, the problem can be solved based on a subspace containing the sub-solutions. Specifically, we add a new variable in the optimization problem, which is $\hat{\mathbf{g}}$ representing the mapping of $\hat{\mathbf{h}}$ in a specific subspace: $\hat{\mathbf{h}} \rightarrow \hat{\mathbf{g}} \in \mathcal{S}$. The goal is to explicitly impose the subspace constraints by the cloned variables, though this will inevitably bring extra storage costs due to the replicated variables. We have a new  optimization problem as:
		
\begin{equation}
	\begin{array}{l}
		\textit{minimize} \quad \textit{E}(\hat{\mathbf{h}}) \\
		\textit{subject to} \quad \hat{\mathbf{h}} = \hat{\mathbf{g}}; \hat{\mathbf{g}} \in \mathcal{S},
		\end{array}
	\label{eq:SADMM1}
\end{equation}
where $\mathcal{S}$ refers to a well-designed subspace. ALMs are used to solve the problem via
\begin{equation}
	\begin{array}{l}
		E_S(\hat{\bf h}) = 	E(\hat{\bf h}) + {{R}}^T (\hat{\bf g} - \hat{\bf h})+  \frac{\sigma}{2} ||\hat{\bf g} - \hat{\bf h}|{|^2},  \\
				
	\end{array}
	\label{eq:SADMM2}
\end{equation}		
where ${R}^T$ denotes the Lagrange multiplier vector, and  $\sigma$ is a regularization term. Based on the classical ADMM method which uses partial updates for the dual variables, $\hat{\bf g}, \hat{\bf h}  $ are then solved as follows:
		
\begin{equation}
	\begin{array}{l}
	{{\hat{\bf h}}^{t + 1}} = \operatorname{argmin} \textit{E}_S({{\hat{\bf h}}|{\hat{\bf g}}^{t}})
	\end{array}
	\label{eq:SADMM3}
\end{equation}
		
\begin{equation}
	\begin{array}{l}
	{{\hat{\bf g}}^{t + 1}} = \operatorname{argmin} \textit{E}_S({{\hat{\bf g}}|{\hat{\bf h}}^{t+1}}) \\
	\textit{subject to} \quad \hat{\mathbf{g}} \in \mathcal{S}.
	\end{array}
	\label{eq:SADMM4}
\end{equation}
Different from Eq.~(\ref{eq:SADMM3}) that can be easily solved using an existing optimization method, i.e., Gradient descent, the solution of Eq.(\ref{eq:SADMM4}) becomes complex due to the new constraint of $\mathbf{g} \in \mathcal{S}$. We rewrite Eq.~(\ref{eq:SADMM4}) by dropping the index for an easy presentation:
\begin{equation}
	\begin{array}{l}
	{{\hat{\bf g}}} = \operatorname{argmin}
	 {{R}}^T (\hat{\bf g} - \hat{\bf h})+  \frac{\sigma}{2} ||\hat{\bf g} - \hat{\bf h}|{|^2},  \\
	\textit{subject to} \quad \hat{\mathbf{g}} \in \mathcal{S}.
	\end{array}
	\label{eq:SADMM5}
\end{equation}
By dropping constant terms, Eq.(\ref{eq:SADMM5}) is equivalent to:
\begin{equation}
 	\begin{array}{l}
 	{{\hat{\bf g}}^{}} = \operatorname{argmin}
 	
 	||\hat{\bf g} - ( \hat{\bf h} - \frac{R}{\sigma})|{|^2},  \\
 	\textit{subject to} \quad \hat{\mathbf{g}} \in \mathcal{S}.
 	\end{array}
 	\label{eq:SADMM6}
\end{equation}
 	 	
The solution to Eq.(\ref{eq:SADMM6}) is given by:
$\hat{\bf g} = \mathbf{M}_s ( \hat{\bf h} - \frac{{\color{blue}{r}}}{\sigma})$, where $\mathbf{M}_s$ is the projection matrix related to the subspace of the solutions to our problem. $R$ is omitted to obtain the ADMM scheme from the original ALM method \cite{mct}. This also improves the efficiency of our method. Thus, we have:
\begin{equation}
	\begin{array}{l}
	E_S(\hat{\bf h}) = 	E(\hat{\bf h}) +  \frac{\sigma}{2} ||\hat{\bf g} - \hat{\bf h}|{|^2}. \\
		
	\end{array}
	\label{eq:SADMM7}
\end{equation}
		
In the relaxed version $ \hat{\bf g}$ is only considered to be recovered by $\mathcal{S}$ ($\mathbf{M}_s$) built from   $\hat{\bf h}$, which is defined by the function $\Phi$ discussed later. This leads to a new ADMM method, named subspace based ADMM (SADMM) algorithm, which makes use of an iterative process similar to that in \cite{admm}.
Specifically, after the variable replication, $\hat{\mathbf{g}}$ is calculated according to a given subspace. This means that we could find  $\hat{\mathbf{h}}^{t+1}$ based on $\hat{\mathbf{h}}^{t}$ and $\hat{\mathbf{g}}^{t}$ in the $t^{th}$ iteration. Next, we expand the training set by adding a number of training samples.  $\hat{\mathbf{g}}^{t+1}$ is calculated based on the subspace spanned by $\hat{\mathbf{h}}^{0:t+1}$ (($\mathbf{M}_s$)), which includes sub-filters from $\hat{\mathbf{h}}^{0}$ (initialized) to $\hat{\mathbf{h}}^{t+1}$. This iterative process is described as follows:
\begin{equation}
	\begin{array}{l}
	{{\hat{\bf h}}^{t + 1}} = \operatorname{argmin} \textit{E}_S({{\hat{\bf h}}|{\hat{\bf g}}^{t}}),\\
	{{\hat{\bf g}}^{t + 1}} = \Phi ({{\hat{\bf h}}^{t + 1}}, {{\hat{\bf h}}^{0:t}}).
	\end{array}
	\label{eq:SADMM8}
\end{equation}
It should be noted that the theoretical investigation into our SADMM algoritm shows that the convergence speed of SADMM is as fast as   ADMM \cite{admm}, which is elaborated in the appendix part.

\section{Latent constrained correlation filters}
\label{sec:lccf}
		
\subsection{Correlation filters}
\label{subsec:mccf}
The solution to correlation filters, i.e., multiple-channel correlation filter, can be regarded as an optimization problem which minimizes $\rm{E}(\bf{h})$. This procedure can be described by the following objective function:
\begin{equation}
	E({\bf{h}}) = \frac{1}{2}\sum\limits_{i = 1}^N {||{{\bf{y}}_i} - \sum\limits_{k = 1}^K {{{\bf{h}}_{[k]}^{}}^T \otimes {{\bf{x}}_{i,[k]}}} ||_2^2}  + \frac{1}{2}\sum\limits_{k = 1}^K {||{{\bf{h}}_{[k]}^{}}||_2^2},
	\label{eq:MCCF1}
\end{equation}
where $N$ represents the number of images in the training set, and $K$ is the total number of channels.
In Eq.~(\ref{eq:MCCF1}), $\mathbf{x}_i$ refers to a K-channel feature/image input which is obtained in a texture feature extraction process, while $\mathbf{y}_i$ is the given response whose peak is located at the target of interest. $k$ represents the $k^{th}$-channel. The single-channel response is a $D$ dimensional vector, $\mathbf{y}_i = [ \mathbf{y}(1),...,\mathbf{y}(D)]^T \in \mathbb{R}^D$. Both $\mathbf{x}_i$ and $\mathbf{h}$ are $K \times D$ dimensional super-vectors that refer to multi-channel image and filter, respectively. Correlation filters are usually constructed in the frequency domain. Therefore, to solve Eq.~(\ref{eq:MCCF1}) efficiently, we transform the original problem into the frequency domain by the Fast Fourier Transform (FFT), which becomes:	
\begin{equation}
	E({\bf{\hat h}}) = \frac{1}{2}\sum\limits_{i = 1}^N {||{{{\bf{\hat y}}}_i} - \sum\limits_{k = 1}^K {{\rm{diag}}{{({{{\bf{\hat x}}}_{i,[k]}})}^T}
	{{{\bf{\hat h}}}_{[k]}^{}}} } ||_2^2 + \frac{\lambda }{2}\sum\limits_{k = 1}^K {||{{{\bf{\hat h}}}_{[k]}^{}}||_2^2},
	\label{eq:MCCF2}
\end{equation}
where $\hat{\mathbf{h}}$, $\hat{\mathbf{x}}$, and $\hat{\mathbf{y}}$ refer to the Fourier form of $\mathbf{h}$, $\mathbf{x}$, and $\mathbf{y}$, respectively. Eq.~(\ref{eq:MCCF2}) can be further simplified to:
\begin{equation}
	E_L({\bf{\hat h}}) = \frac{1}{2}\sum\limits_{i = 1}^N {||{{{\bf{\hat y}}}_i} - {{{\bf{\hat X}}}_i}} {\bf{\hat h}}||_2^2 + \frac{\lambda }{2}||{\bf{\hat h}}||_2^2,
	\label{eq:MCCF3}
\end{equation}
where 	
\begin{equation}
\begin{array}{l}
{\bf{\hat h}} = {[{{{\bf{\hat h}}}_{[1]}^{T}},...,{{{\bf{\hat h}}}_{[k]}^{T}}]^T}\\
{{\bf{\hat X}}_i} = [{\rm{diag}}{({{\bf{{\hat x}}}_{i,[1]}})},...,{\rm{diag}}{({{\bf{{\hat x}}}_{i,[k]}})}]
\end{array}
\label{eq:MCCF4}
\end{equation}
A solution in the frequency domain is given by:
\begin{equation}
	\hat{\bf h} = {(\lambda {\bf{I}} + \sum\limits_{i = 1}^N {{{{\bf{\hat X}}}_i}^T{{{\bf{\hat X}}}_i}})^{ - 1}}\sum\limits_{i = 1}^N {{{{\bf{\hat X}}}_i}^T{{{\bf{\hat y}}}_i}}.
	\label{eq:MCCF5}
\end{equation}
Here, since $\mathbf{X}_i$ is usually a sparse banded matrix, one can actually transform solving the $KD \times KD$ linear system into solving $D$ independent $K \times K$ dimensional linear systems. By doing so, the correlation filters calculation exhibits excellent computational and memory efficiency.

\subsection{Latent constrained linear correlation filter \\(LC-LCF) based on SADMM}
\label{sec:proposed}
	
In order to solve Eq.~(\ref{eq:MCCF3}) based on SADMM, we reformulate it using the subspace constraint as:
\begin{equation}
	\begin{array}{l}
		\textit{minimize} \quad \textit{E}_L(\hat{\mathbf{h}}) \\
		\textit{subject to} \quad \hat{\mathbf{h}} = \hat{\mathbf{g}}; \hat{\mathbf{g}} \in \mathcal{S}.
		\end{array}
		\label{eq:LCLCF1}
\end{equation}
The objective function in Eq.~\eqref{eq:LCLCF1} can be expressed as:
	\begin{equation}
		\begin{array}{l}
			E_{S,L}(\hat{\bf h}) = \frac{1}{2}\sum\limits_{i = 1}^B {||{{\hat{\bf y}}_i} - {{\hat{\bf X}}_i} \hat{\bf h}} ||_2^2 + \frac{\lambda }{2}||\hat{\bf h}||_2^2 +  \frac{\sigma}{2} ||\hat{\bf g} - \hat{\bf h}|{|^2},  \\			
		\end{array}
		\label{eq:LCLCF2}
	\end{equation}
where $\lambda$ and $\sigma$ are regularization terms. 
According to SADMM, the solution is described as follows:
	\begin{equation}
		\begin{array}{l}
			{{\hat{\bf h}}^{t + 1}} = \operatorname{argmin} \textit{E}_{L,S}({{\hat{\bf h}}|{\hat{\bf g}}^{t}}),\\
			{{\hat{\bf g}}^{t + 1}} = \Phi ({{\hat{\bf h}}^{t + 1}}, {{\hat{\bf h}}^{0:t}}).
		\end{array}
		\label{eq:LCLCF3}
	\end{equation}
	Here $\mathbf{M}_s$ is defined to be ${\hat{\bf h}}^{0:t}$.
	To solve Eq.(\ref{eq:LCLCF3}), we calculate the partial derivatives of Eq.~(\ref{eq:LCLCF2}), and thus have:
	\begin{equation}
    \begin{aligned}
	\frac{\partial{E_{S,L}(\hat{\bf h}^{t+1})}}{\partial{(\hat{\bf h}^{t+1})}}
	&	= \sum_{i=1}^B {({{\hat{\bf X}}_i}^T{{\hat{\bf X}}_i} + \lambda \mathbf{I} + \sigma \mathbf{I}) \hat{\bf h}^{t+1}} \\
    &- \sum {{{{\bf{\hat X}}}_i}^T{{{\bf{\hat Y}}}_i} - \sigma {{{\bf{\hat g}}}^{t}}},
	\label{eq:LCLCF4}
    \end{aligned}
	\end{equation}	
	\begin{algorithm}[!t]
		\caption{LC-LCF based on SADMM} \label{alg:sc_mccf}
		\begin{algorithmic}[1]
			\State Set $k=0$, $\varepsilon_{best} = + \infty$, $\eta = 0.7$
			\State Initialize $\sigma^{0} = 0.25$ (suggested in \cite{mct})
			\State Initialize $\hat{\mathbf{g}}^{0}$ and $\hat{\mathbf{h}}^{0}$ based on MCCF
			\State Initialize $B$,  $B$ denotes the size of half of training samples, \textit{maxiter} = 12
			\Repeat
			\State ${\bf{H}} = \sum\limits_{i = 1}^B {({{\hat{\bf X}}_i}^T{{\hat{\bf X}}_i})}  + \lambda \mathbf{I} + {\sigma}^{t} \mathbf{I}$
			\State ${\hat{\bf h}^{t + 1}} = {{\bf{H}}^{-1}} \left(\sum\limits_{i = 1}^B {{{\hat{\bf X}}_i}^T{{\hat{\bf Y}}_i}}  + {\sigma}{\hat{\bf g}^{t}} \right)$
			\State $\varepsilon  = \| \hat{\mathbf{h}}^{t + 1} - \hat{\mathbf{h}}^{t} \|_2$
			\If{$\varepsilon  < \eta \times \varepsilon_{best}$}
			\State $\sigma^{t+1} = \sigma^{t}$
			\State $\varepsilon_{best} = \varepsilon$
			\Else
			\State $\sigma^{t+1} = 2 \sigma^{t}$
			\EndIf
			\State ${\hat{\bf g}^{t + 1}} = \Phi(\hat{\mathbf{h}}^{t+1}, \hat{\mathbf{h}}^{0:t}) $
			
			\State $t \leftarrow t+1$, $B \leftarrow B + \textit{B}/\textit{maxiter}$
			\Until{some stopping criterion, i.e., maximum number of iteration (\textit{maxiter}=12). }
		\end{algorithmic}
	\end{algorithm}
where $B$ is the size of the training set. We come to the result of $\hat{\mathbf{h}}^{t+1}$, and have:
	\begin{equation}
		{\hat{\bf h}^{t + 1}} = {{\bf{H}}^{-1}}(\sum\limits_{i = 1}^B {{{\hat{\bf X}}_i}^T{{\hat{\bf Y}}_i}}  +  {\sigma ^{t}}{\hat{\bf g}^{t}}),
		\label{eq:LCLCF5}
	\end{equation}
where
	\begin{equation}
		{\bf{H}} = \sum\limits_{i = 1}^B {({{\hat{\bf X}}_i}^T{{\hat{\bf X}}_i})}  + \lambda \mathbf{I} + {\sigma ^{t}} \mathbf{I},
		\label{eq:LCLCF6}
	\end{equation}
then $\hat{\mathbf{g}}^{t+1}$ is calculated as :
	\begin{equation}
		{\hat{\bf g}^{t + 1}} = \Phi(\hat{\mathbf{h}}^{t+1}, \hat{\mathbf{h}}^{0:t}) =
		\sum\limits_{i = 0}^t {\omega_i}\hat{\mathbf{h}}^{i},
		\label{eq:LCLCF7}
	\end{equation}
where $\omega_i=\frac{1}{d_i}$, and
$d_i$ is the Euclidean distance between ${\hat{\bf h}}^{t + 1}$ and ${\hat{\bf h}}^{i}$. $\omega$ will be normalized by the $L$1 norm.After several iterations, $\hat{\mathbf{h}}^{t+1}$ converges to a saddle point, which is proved in Appendix I. The pseudocode of our proposed method is summarized in Algorithm~\ref{alg:sc_mccf}. The LC-LCF is first initialized based on a half of the training samples, and then we add $\frac{B}{maxiter}$ samples into the training set, which is one kind of data sampling. Subsequently, a set of sub-filters (solution sampling) are calculated, and further used to constrain our final solution.

\begin{algorithm} \caption{\textbf{- LC-KCF algorithm for object tracking}}
	\label{alg:proposed tracking algorithm}
	\begin{algorithmic}[1]
		\State  Initial target bounding box ${\bf{b}}_1=[p_x,p_y,w,h] $,
		\State  Initial ${\hat{\bm\alpha}}^0$ using KCF method,${\hat{\bm\beta}}^0={\hat{\bm\alpha}}^0$,${\hat{\bm\sigma}}^0=0.0001$
		\State  Initial ${\epsilon}_{best}=\infty$,$c=2$,$\lambda=0.0001$,$t=1$,
		\State {Initial $T$}
		\Repeat
		
		\State Crop out the search windows  according to $\bf{b}_{t}$, and extract the HOG features for training
		\State Compute the kernel matrix ${\bf K}^{xx}$
		\State Update ${\bm \eta}$ using Eq.~\eqref{eq:LCKCF7}
		\State Update ${\hat{\bm \alpha}}^{t+1}$ using Eq.~\eqref{eq:LCKCF8}
		\If{$t<=T$}
		\State Update 
		${\hat{\bm\beta}}^{t+1}=\sum_{i=0}^t{{\bm\omega}_i}{\hat{\bm\alpha}}^i$
		\Else
		\State Update ${\hat{\bm\beta}}^{t+1}=\sum_{i=t-T+1}^t{{\bm\omega}_i}{\hat{\bm\alpha}}^i$
		\EndIf
		\State $\epsilon={||{\hat{\bm\alpha}}^{t+1}-{\hat{\bm\alpha}}^{t}||}$
		\If{$\epsilon<{\epsilon}_{best}$}
		\State $\sigma^{t+1}={\sigma}^t$,${\epsilon}_{best}=\epsilon$
		\Else
		\State $\sigma^{t+1}=c{\sigma}^t$
		\EndIf
		\State  Crop out the search window  and extract the HOG features for testing
		\State  Compute the kernel matrix of the test frame ${\bf K}^{zx}$
		\State  Compute the correlation response as:${\hat y}=\mathcal{F}^{-1}({\bf K}^{zx}\odot{\hat{\bm \alpha}}^{t+1})$
		\State  Select the coordinate of the maximal correlation response as the next location of tracking object
		\State  Update ${\bf b}_{t+1}$
		\State  $t\leftarrow t+1$
		\Until{End of the video sequence.}
		\State \textbf{end}
	\end{algorithmic}
\end{algorithm}


\subsection{Latent constrained kernelized correlation filter (LC-KCF) based on SADMM}
\label{LC-KCF}
In kernelized correlation filter (KCF), similar to the linear case, filters learned on the previous frames can also be used to constrain the solution. Details about KCF can refer to Appendix II. If one frame is disturbed by occlusion and noise, the performance of the filter tends to drop. Our idea is that, with a subspace constraint, the filter is regularized by a projection into a well-defined subspace to achieve a higher performance. In other words, the samples of the previous frames are involved in the reconstruction process of the filter with different weights, therefore enhancing the robustness of the filters.

As described in the second KKT condition as shown in Eq.~(\ref{eq:KCF4}) (Appendix II), we can solve ${\bf h}$ in a dual space by setting $\bm\alpha=\frac{\bm\theta}{2\lambda}$. For KCF, the latent constraint is actually made for $\bm\alpha$. We introduce the constraint term $\bm\beta$ as the mapping of $\bm{\alpha}$ in the  subspace: (${{\bm\alpha}}\rightarrow{{\bm\beta}},{{\bm\beta}}\in \mathcal{S}$). Therefore, Eq.~\eqref{eq:KCF2} can be rewritten as:
\begin{equation}
	\begin{aligned}
		\mathcal{L}_p=&\sum_{i=1}^{M\times N}{{{\xi}}_i^{t+1}}^2+\sum_{i=1}^{M\times N}{{\theta}_i^{t+1}({y}_i-({\bf h}^{t+1})^T{\bm\phi}_i-{\xi}_i^{t+1})}\\
		&+\lambda(||{\bf h}^{t+1}||^2-C^2)-\delta^{t}||{{\bm\alpha}^{t+1} -{\bm\beta}^t}||_2^2.
	\label{eq:LCKCF1}
	\end{aligned}
\end{equation}
The superscript $t$ refers to the frame index, and also denotes the iteration number.
The solution to Eq.~\eqref{eq:LCKCF1} based on the SADMM method also depends on the iterative process. The variables need to be updated include ${\bm\alpha}$, ${\bm\beta}$, and punishment coefficient $\delta$. The mapping function $\bm\phi_i = \bm\phi(x_i)$, which is used to calculate the kernel matrix  $\bf K$. 
Based on KCF, whose details are introduced in the Appdenxi part, we come up with a new maximizing objective function $\textit{E}(\bm\alpha)$:

\begin{equation}
	\begin{aligned}
		\textit{E}(\bm \alpha)=&-\lambda^2\sum_{i=1}^{M\times N}{\left({\alpha}_i^{t+1}\right) ^2}+2\lambda\sum_{i=1}^{M\times N}{{\alpha}_i^{t+1}{y}_i}\\
		&-\lambda\sum_{i,j=1}^{M\times N}{{\alpha}_i^{t+1}{{\alpha}}_j^{t+1}}{{\bf K}_{i,j}}-\delta^{t}\sum_i^{M\times N}{({\alpha}_i^{t+1}-{\beta}_i^t)^2}\\
		&-\lambda C^2.
		\label{eq:LCKCF2}
	\end{aligned}
\end{equation}
We can further simplify it as:
\begin{equation}
	\begin{aligned}
		{E}({\bm\alpha})
        =&-\lambda^2{({\bm\alpha}^{t+1})}^T{({\bm\alpha}^{t+1})}+2\lambda{({{\bm\alpha}^{t+1}})^T{{\bf y}}}\\ &-\lambda{{({\bm\alpha}^{t+1})}^T{\bf K}{\bm\alpha}^{t+1}}\\
		&-\delta^t({{\bm\alpha}}^{t+1}-{{\bm\beta}}^t)^T({{\bm\alpha}}^{t+1}-{{\bm\beta}}^t) -\lambda B^2.
		\label{eq:LCKCF3}
	\end{aligned}
\end{equation}
By setting a variable substitution $\delta^t=\lambda\sigma^t$ and taking derivatives w.r.t. $(\bm\alpha^{t+1})$:
\begin{equation}
    \begin{aligned}
	\frac{{\partial {({\bf{E}(\bm\alpha)})}}}{{\partial ({{{{\bm\alpha}^{t+1}}}})}} = &-2\lambda^2{{\bm\alpha}^{t+1}}+2\lambda{\bf y}-2\lambda{{{\bf K}{\bm\alpha}}^{t+1}}\\
	&-2\lambda\sigma^t({\bm\alpha}^{t+1}-{\bm\beta}^t),
	\end{aligned}
	\label{eq:LCKCF4}
\end{equation}
we come to the solution of ${\bm\alpha}^{t+1}$ as:
\begin{equation}
	{\bm\alpha}^{t+1}=({{\bf K}+\lambda {\bf I}+\sigma^t {\bf I}})^{-1}({{{\bf y}}+\sigma^t{{\bm \beta}}^t}).
	\label{eq:LCKCF5}
\end{equation}

The kernel matrix $\bf K$ is a cyclic matrix, we transform Eq.~(\ref{eq:LCKCF5}) by Fast Fourier Transform (FFT) to avoid the inverse operation of matrices \cite{kcf}:
\begin{equation}
{\bf K}={F}^H{\bf K}^{xx}{F},
\end{equation}
where $F$ is a discrete Fourier transform matrix, and ${F}^H$ is the conjugate transpose operation. ${\bf K}^{xx}$ is the first row of ${\bf K}$, then we obtain
\begin{equation}
	{\hat{\bm\alpha}}^{t+1}=({{\bf K}^{xx}+\lambda+\sigma^t})^{-1}({{\hat{\bf y}}+\sigma^t{\hat{\bm \beta}}^t}).
	\label{eq:LCKCF6}
\end{equation}
By setting
\begin{equation}
	{\bm\eta}=({{\bf K}^{xx}+\lambda {\bf I}+\sigma^t {\bf I}})^{-1}({{\bf K}^{xx}+\lambda I}),
	\label{eq:LCKCF7}
\end{equation}
we change Eq.~\eqref{eq:LCKCF6} to another form
\begin{equation}
{\hat{\bm\alpha}}^{t+1}={\bm\eta}{\hat{\bm\alpha}}+\left(1-\bm\eta\right){\hat{\bm\beta}}^t.
\label{eq:LCKCF8}
\end{equation}
This means that for the $t^{th}$ frame, we can calculate ${\hat{\bm \alpha}}^{t+1}$ based on ${\hat{\bm\beta}}^t$, and then update ${\hat{\bm\beta}}^{t+1}$ based on the subspace spanned by ${\hat{\bm \alpha}}^{0:t+1}$ as the linear case. The process mentioned above can be summarized as:
\begin{equation}
	\left\{
	\begin{aligned}
		&{\hat{\bm\alpha}}^{t+1}={\bm\eta}{\hat{\bm\alpha}}+\left(1-\bm\eta\right){\hat{\bm\beta}}^t\\
		&{\hat{\bm\beta}}^{t+1}= \Phi(\hat{{\bm\alpha}}^{t+1}, \hat{{\bm\alpha}}^{0:t})  = \sum_{i}{\bm\omega_i} {\hat{\bm\alpha}}^i\\
		&{\hat{\sigma}}^{t+1}=c{\hat{\sigma}}^t
	\end{aligned}
	\right.
	\label{eq:LCKCF9}
\end{equation}
Similar to Eq. (\ref{eq:LCLCF7}), $\omega_i$ is calculated based on the Euclidean distance and normalized by the $L$1 norm, and details refer to the source code. In our experiment, we set $c = 2$. 

\section{Experiments}
\label{sec:experiment}
\begin{figure}[!t]
	\centering
	\includegraphics[width=0.5\textwidth]{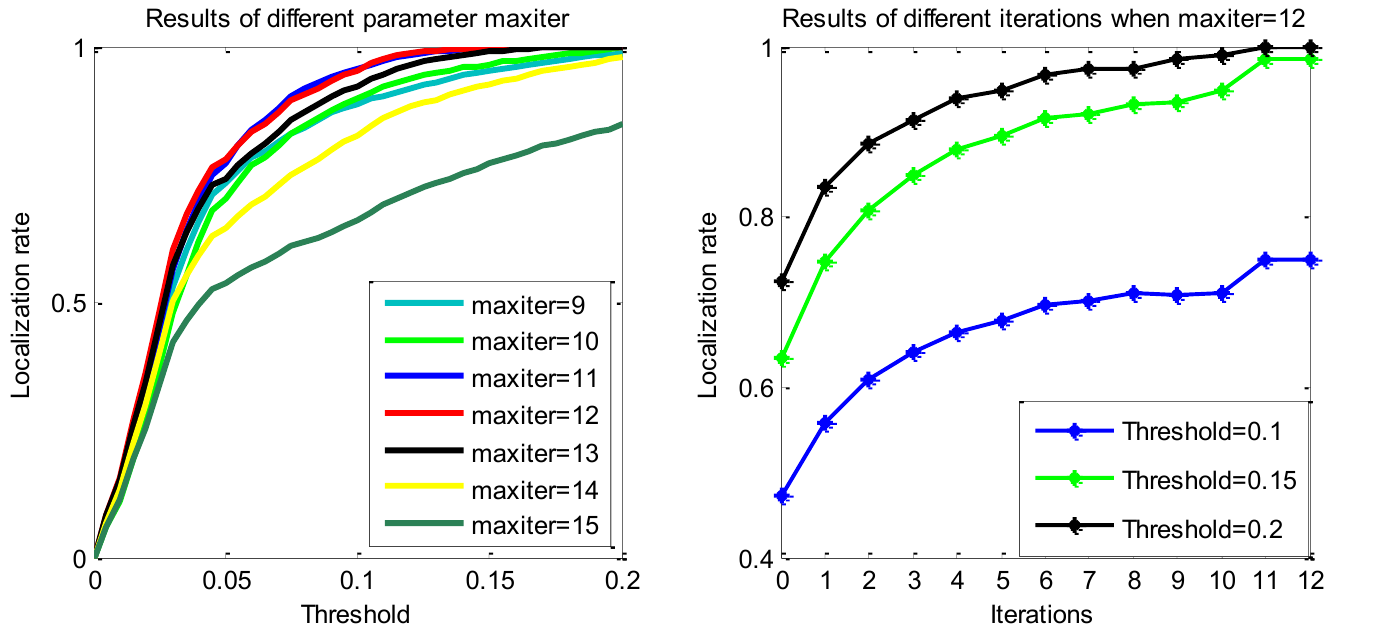}
	\caption{Left: the localization rates under different maximum iterations for LCCF. Right: the convergence of our method when fixing the $ maxiter$ to 12.}
	\label{fig:iter}
\end{figure}
\begin{figure}[!h]
	\centering
	\includegraphics[width=8cm]{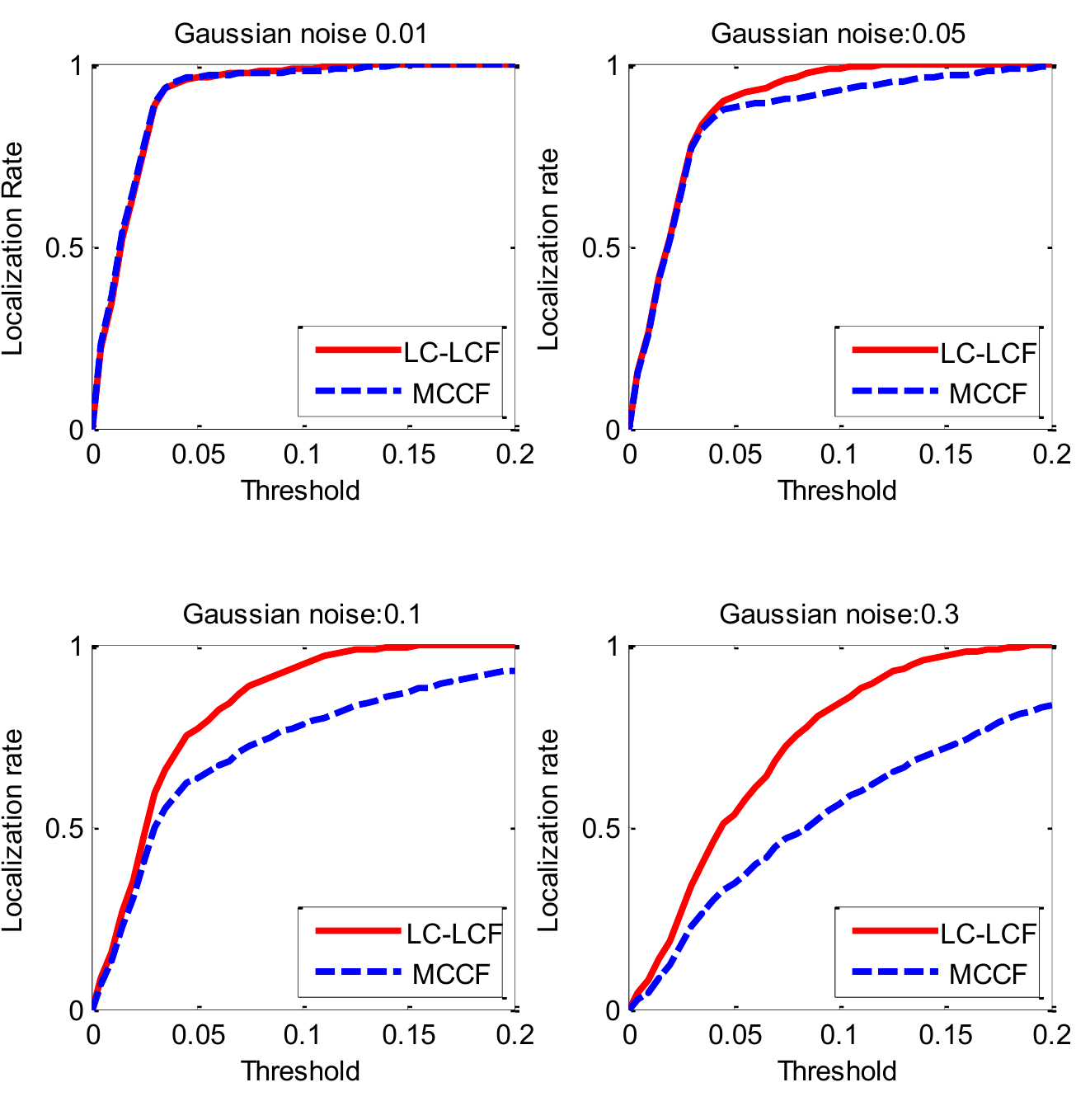}
	\caption{The comparison between LCCF and MCCF on CMU Multi-PIE with different Gaussian noise parameters}
	\label{fig:cmu}
\end{figure}
	
\label{sec:experiment}
In this section, to evaluate the performance of the proposed method, experiments are carried out for object detection and tracking. For objection detection, two different applications are considered: eye localization and car detection. A two-dimensional Gaussian function with the same parameter is employed to generate a single channel output whose peak is located at the coordinate of target. All images are normalized before training and testing. The images are power normalized to have a zero-mean and a standard deviation of 1.0. 
	
\textbf{Subset, subspace and robustness evaluation for object detection.} We first introduce how to create different kinds of subsets for calculating the sub-filers subspace. We add some noise or occlusions to the training and test sets in order to show how LC-LCF can gain robustness by a projection onto a subspace. More specifically, we first select an initial subset containing half of the training samples (the size was denoted by $B$). Other subsets are built by adding $\frac{B}{maxiter}$ samples into the initial subset, where $maxiter$ represents the maximum number of iterations. Based on the initial subset, we obtain $\hat{\mathbf{h}}^{[0]}$, and calculate other sub-filters, i.e.,  $\hat{\mathbf{h}}^{t}$ for the $t^{th}$ iteration, step by step. With respect to the robustness evaluation, the basic idea for both applications is to measure the algorithm accuracy when adding Gaussian noise or occlusions to the training and test sets. For both applications, HOG feature is extracted by setting the number of direction gradients to 5, and both the sizes of block and cell to [5,5], as suggested in \cite{mccf}.

\subsection{Eye localization}
\label{sec:eye}
In the first experiment, the proposed method is evaluated for eye localization and compared with several state-of-the-art correlation filters, including MCCF \cite{mccf}, correlation filters with limited boundary (CFwLB) \cite{boundary}, ASEF \cite{asef} and MOSSE \cite{mosse}.

\subsubsection{CMU Multi-PIE face database}
The CMU Multi-PIE face database is used in this experiment. It consists of 902 frontal faces with neutral expression and normal illumination. We randomly select 500 images for training and the remaining for testing. All images are cropped to have a same size of 128 $\times$ 128 with fixed coordinates of the left and right eyes. We train a 128$\times$128 filter of the right eye using full face images by following \cite{mccf}. Similar to ASEF and MOSSE, we define the desired response as a 2D Gaussian function with a spatial variance of 2. Eye localization is performed by correlating the filters over the test images followed by selecting the peak of the output as the predicted eye location.
	
	
\textbf{Results and analysis.} In order to evaluate the performance of our algorithm, we use the so-called fraction of interocular distance, which is defined by the actual and the predicted positions of the eyes. This distance can be computed as
\begin{equation}
	d = \frac{{||{\mathbf{p}_i} - {\mathbf{m}_i}|{|_2}}}{{||{\mathbf{m}_l} - {\mathbf{m}_r}|{|_2}}},
	\label{eq:eye}
\end{equation}
where $\mathbf{p}_i$ is the predicted location by our method, and $\mathbf{m}_i$ is the ground truth of the target of interest, i.e., the eye's coordinates $\mathbf{m}_l$ and $\mathbf{m}_r$.
	
Haven calculated the distance $d$, the next step is to compare it with a threshold $\tau$. If $d < \tau$, the result will be considered as a correct one. We count the correct number under this threshold, and compute the ratio of the correct count to the total number of tests as the localization rate. The localization rates under different $maxiter$s are shown in Fig.~\ref{fig:iter}. We can see that LC-LCF obtains the best accuracy when  $maxiter=12$. Therefore, we use this setting for all the following experiments. In addition, we also test the convergence of our method when $maxiter=12$. It is clear that the performance is monotonically increasing as the incremental iteration numbers, which verifies our proof.

\begin{figure}
	\centering
	\includegraphics[width=0.5\textwidth]{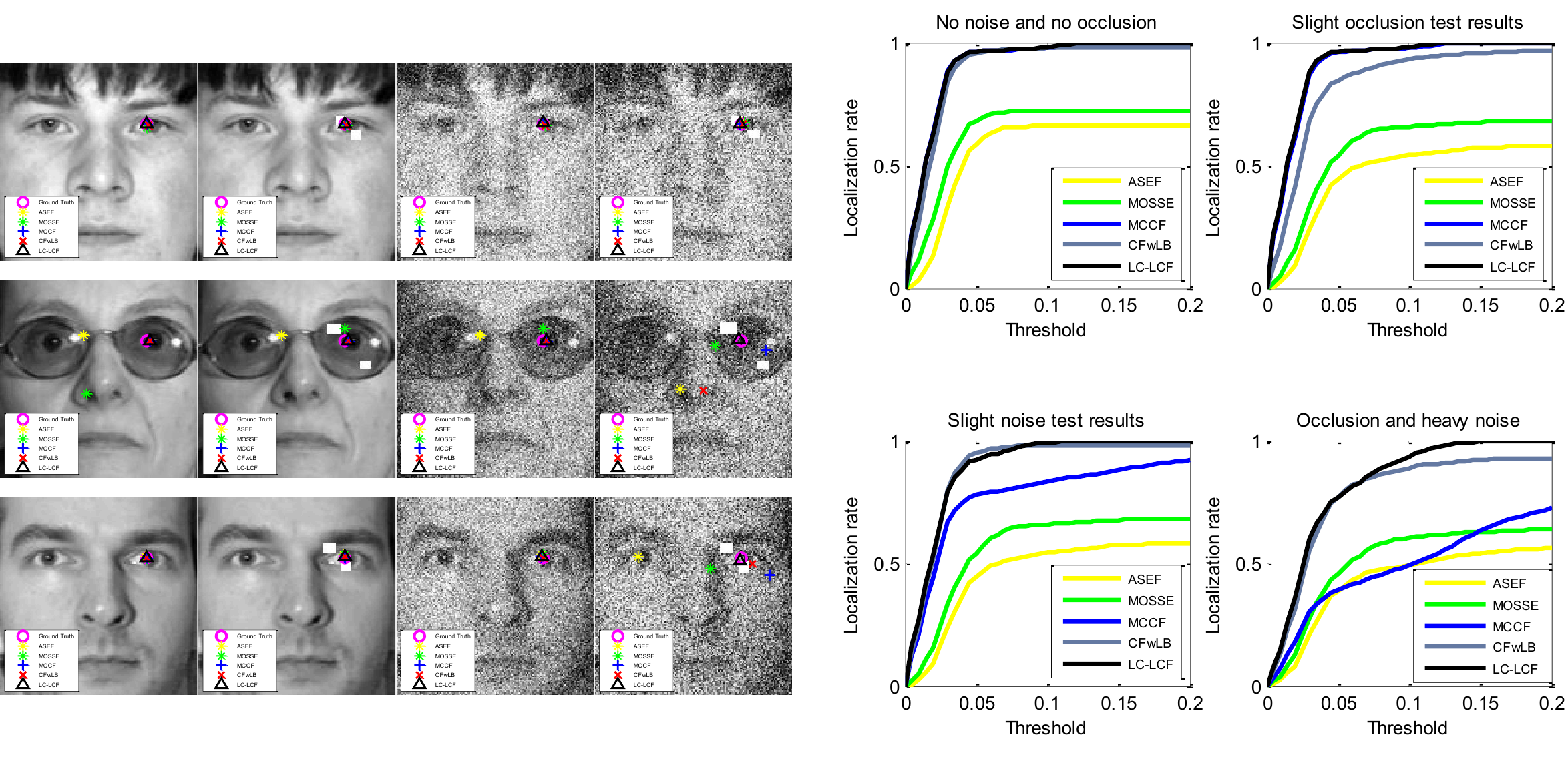}
	\caption{The results of LC-LCF compared to the state-of-the-art correlation filters on CMU Multi-PIE. The variance is varying from 0.05 (slight) to 0.1 (heavy). On the left part, from the first column to the fourth column, we show results on the original images, images with occlusions, images with slight noise, and images with heavy noise and occlusions. }
	\label{fig:cmu1}
\end{figure}

LC-LCF is also compared with MCCF in the robustness evaluation. As shown in Fig.~\ref{fig:cmu}, LC-LCF achieves a much higher performance than MCCF, especially when severe noises are present.  In Fig.~\ref{fig:cmu1}, LC-LCF is compared with the state-of-the-art methods, showing that LC-LCF is less affected by noise and occlusion than others. Particularly, in the situation when the test set is extremely noisy, LC-LCF and CFwLB perform significantly better than other competing approaches. It is also evident that LC-LCF achieves a much better performance than CFwLB for the occlusion case. In these experiments, all methods are based on the same training and test sets. For the experiment on the original dataset  (without noise and occusion), we randomly choose 500 images for training and other 402 for testing. To test the robustness, we further conduct another experiment by adding noise or occlusion onto the selected 500 images, thereby generating a total of 1000 training images. Similarly, the 402 testing images are also added with random noise and occlusion.
This evaluation is repeated for ten times to avoid bias and finally the average accuracy over the ten experiments is reported.

\begin{figure}[!t]
	\centering
	\includegraphics[width=0.5\textwidth]{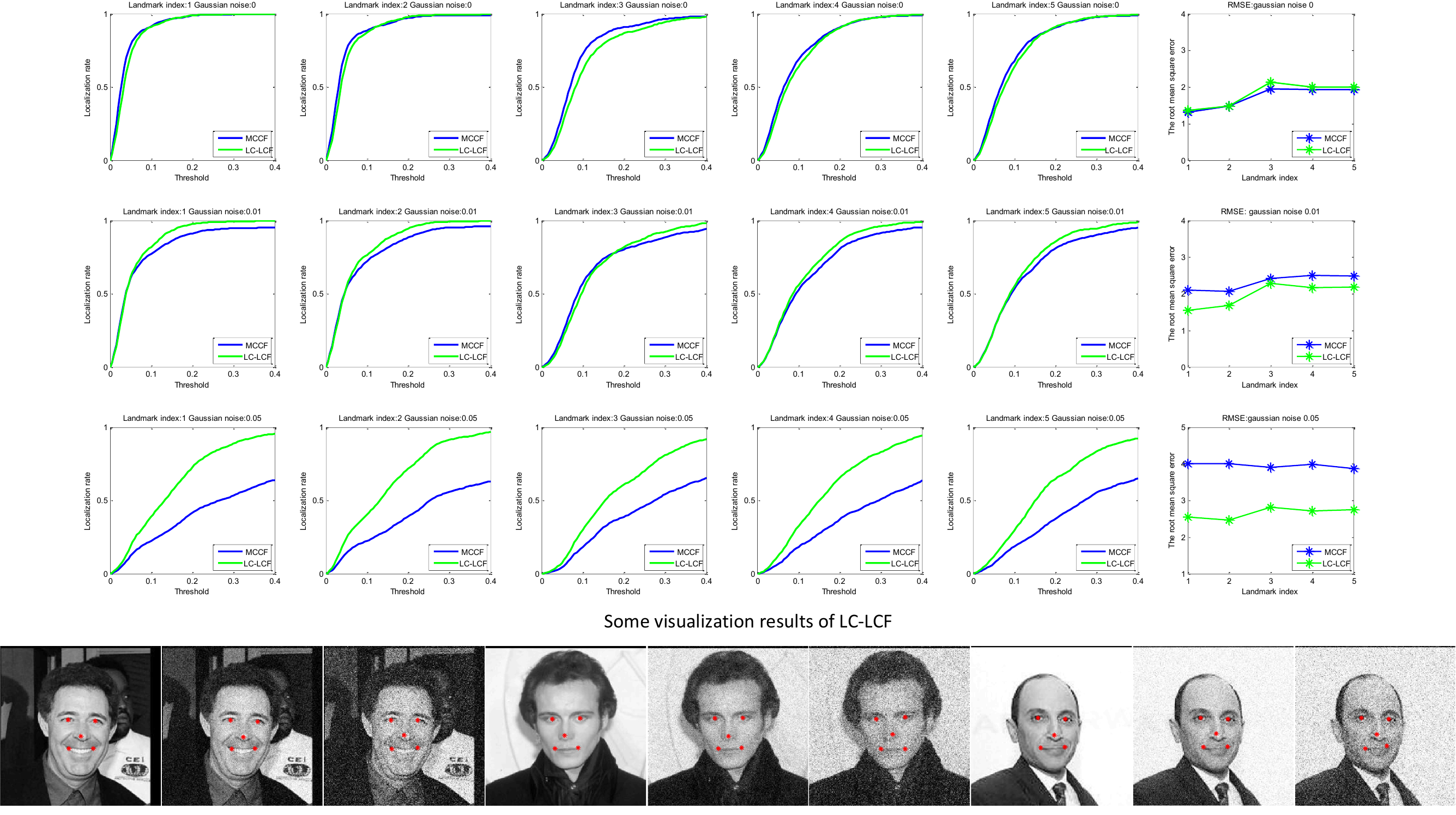}
	\caption{The results of LC-LCF and MCCF on the LFW dataset.}
	\label{fig:lfw}
\end{figure}
	
\subsubsection{LFW database}
In the second eye localization experiment, we choose face images in the Labeled Faces in the Wild (LFW) database. LFW database contains ten thousands of face images, covering different age, sex, race of people. The training samples take into account the diversity of lighting, pose, quality, makeup and other factors as well.
	
We randomly choose 1000 face images of $250 \times 250$ pixels, in which the division for training and testing is half:half.  Fig.~\ref{fig:lfw} shows the predominant robustness of the proposed algorithm. Similar to the results on the CMU dataset, the performance difference between the proposed algorithm and the state-of-the-arts is getting larger as  increased intensity of noise. Considering that LC-LCF is implemented based on MCCF, we only compare the two methods on this dataset. We fail to run the CFwLB code on this database, because it requires the facial points  must be at the same positions for all the images. However, on the following Car dataset, we provide comparisons of all methods.
	
\begin{figure}
	\centering
	\includegraphics[width=0.5\textwidth]{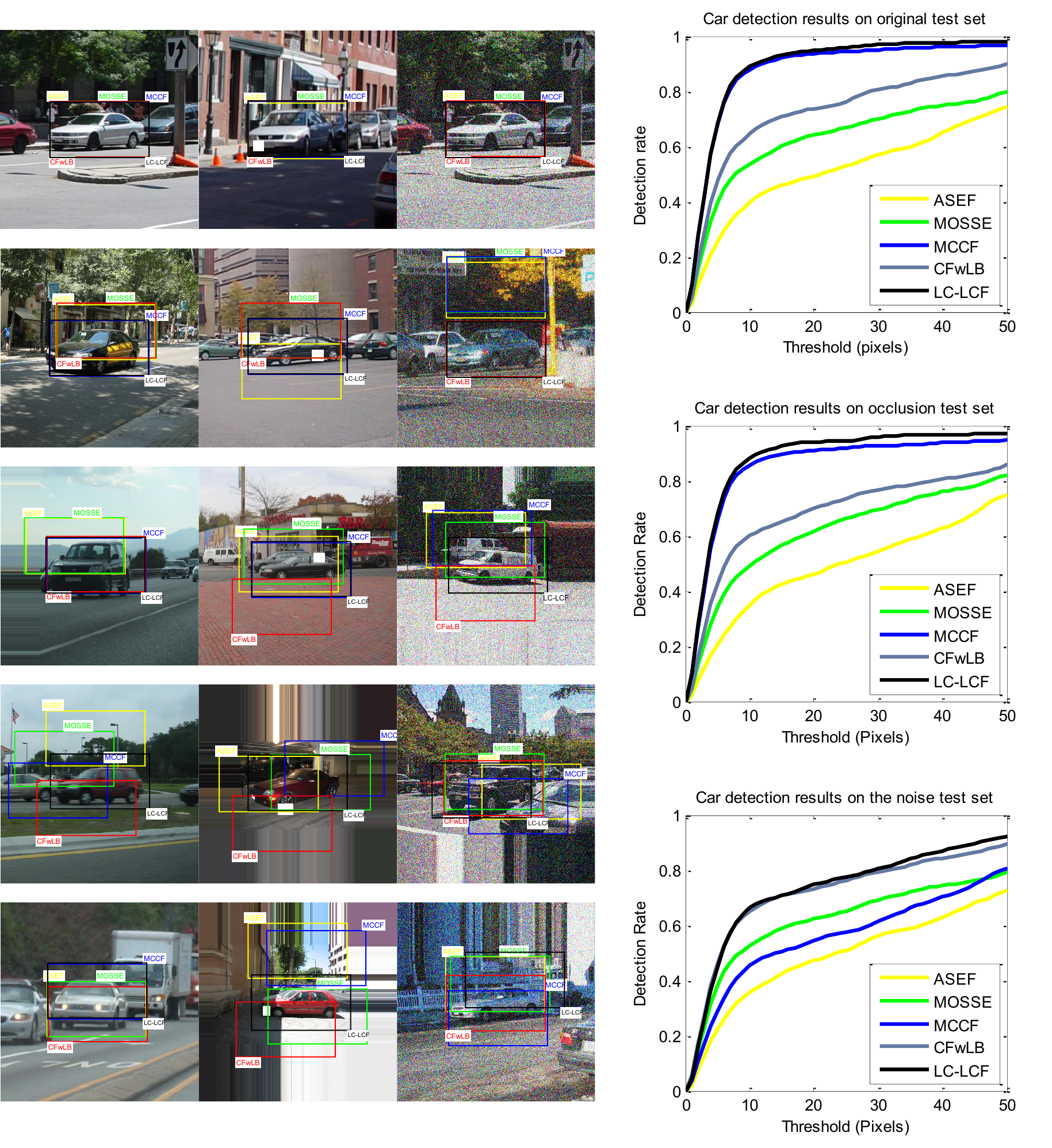}
	\vspace{-0.1cm}
	\caption{Experimental results of LC-LCF compared to other correlation filters for car detection. The variance of Gaussian noise is 0.05. On the left part, from the first column to the third column, we present the results of different methods on the original images, images with occlusion, and images with noise.}
	\vspace{-0.2cm}
	\label{fig:car}
\end{figure}
	
\subsection{Car detection}
\label{sec:car}
	
The car detection task is similar to eye localization. We choose 938 sample images from the MIT Streetscape database \cite{street}. They are cropped to $360 \times 360$ pixels. In the training procedure, HOG feature is used as input and the peak of the required response is located at the center of the car. We use a $100 \times 180$ rectangle to extract the car block and exclude the rest regions in the image.In testing, the peak of the correlation output is selected as the predicted location of a car in the street scene. we compare the predicted location with the  target center, and choose the pixels deviation between them as measurement for evaluation \cite{mccf}. The results of this experiment are presented in Fig.~\ref{fig:car}.


In Fig.~\ref{fig:car}, it can be seen that most methods are quite close to each other in terms of the performance when there is no occlusion or noise. However, LC-LCF shows much better robustness when the test data suffer from noise and occlusion.{The enhanced performance is achieved, because  a subspace that contains various kinds of variations is used to find a more stable and robust solution.}
	
With respect to the complexity, in the testing process, LC-LCF is very fast  since we only need element-wise product in the FFT domain. When we train  $D$ dimensional vector features with $maxiter$ iteration, LC-LCF has a time cost of $\mathcal{O}(N DlogD)$ for FFT calculation (once per image), which is the same to that of MCCF. The memory storage is $\mathcal{O}(maxiter K D)$ for LC-LCF,  and $\mathcal{O}(K^2D)$ for MCCF. Considering that
$maxiter$ is not very big, LC-LCF is quite efficient on training and testing process.
	
\subsection{Object tracking}
\label{sec:tracking}

\begin{figure} [t!]
	\begin{center}
		\includegraphics[width=0.5\textwidth]{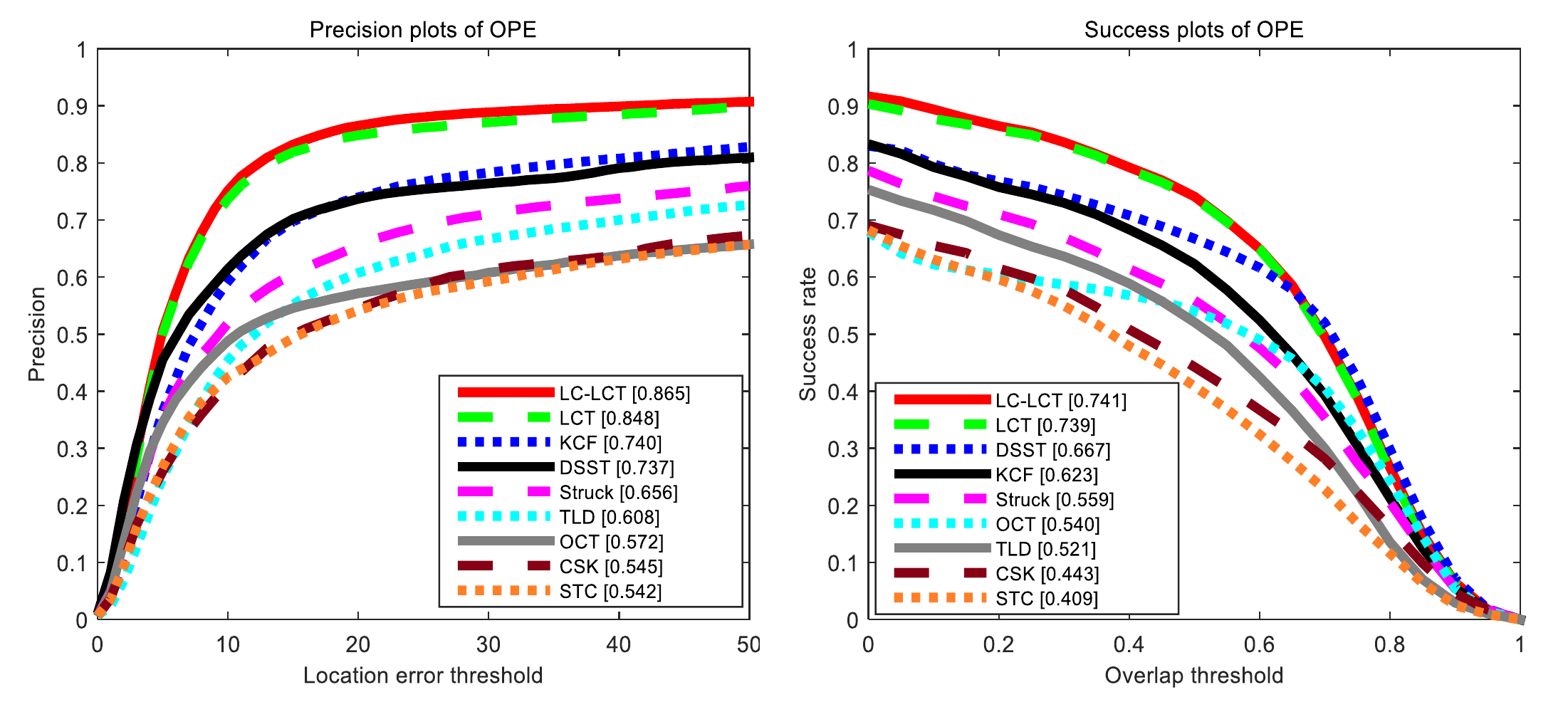}
	\end{center}
	\caption{Success and precision plots according to the online tracking benchmark \cite{benchmark} for long-term tracking experiments.}
	\label{fig:longterm}
\end{figure}

The evaluation of object tracking with the proposed method is conducted on 51 sequences of the commonly used  tracking benchmark \cite{benchmark}. In the tracking benchmark \cite{benchmark}, each sequence is manually tagged with 11 attributes which represent challenging aspects in visual tracking, including \emph{illumination variations}, \emph{scale variations}, \emph{occlusions}, \emph{deformations}, \emph{motion blur}, \emph{abrupt motion}, \emph{in-plane rotation}, \emph{out-of-plane rotation}, \emph{out-of-view}, \emph{background clutters} and \emph{low resolution}. All the tracking experiments are conducted on a computer with an Intel I7 2.4 GZ (4 cores) CPU and 4G RAM. The results show that the tracking performance is significantly improved by adding latent constrains without sacrificing real-time processing. \textbf{The source code will be publicly available}.

\subsubsection{Feature experiments}
To validate the performance of our algorithm on different features, we adopt gray feature, HOG and DCNN feature for comparison. Here, Gaussian kernel function (standard variance = $0.5$) is used. Most parameters utilized in LC-KCF are empirically chosen according to \cite{kcf}: $\lambda =10^{-4}$, $\rho=0.1$, and the searching size is set to 1.5.

In Table.\ref{table:feature}, we report the localization precision, which measures the ratio of successful tracking frames. The criterion of being successfully tracked is that the tracker output is within the certain distance to the ground truth location (typically 20 pixels), measured by the center distance between bounding boxes.
		
Comparing with KCF, our method performs better on these three features. For gray feature, LC-KCF and KCF achieve 56.8\% and 56.1\% localization precision respectively, and LC-KCF achieves a higher overlap success rate (49.5\% vs. 47.3\% ). For the HOG feature, LC-KCF improves the localization accuracy by 5\% (79.4\% vs. 74.0\%). We also compare the performance of our method using deep feature extracted from a VGG-19 model, which is also used in \cite{cfcf}. The results show that the localization precision is improved as well (89.6\% vs 89.1\%) by our method, without bringing in too much additional computational burden. Although the FPS of LC-KCF drops, as compared to KCF, it is still a nearly real-time tracker.

\subsubsection{Parameter experiments}

 In Eq.~(\ref{eq:LCKCF9}),  $T$ represents the number of frames  used to reconstruct the subspace, which has a large impact on performance. For the HOG feature, we set $T = 16 $, which  also appears in Algorithm II. Actually, the value of $T$ is flexible, which can vary from $10$ to $25$ on different trackers or different features. We verify the sensitivity of this parameter by using HOG feature in this section. For HOG features, $T$ is relatively stable, and the results are improved when $T$ is within the range of 12 to 22.

\subsubsection{Long-term tracking experiments}

The tracking targets  may undergo significant appearance variations caused by deformation, abrupt motion, heavy occlusion and out-of-view, which affect the tracking performance significantly. Long-term correlation tracking~\cite{longterm} (LCT) regards tracking as a long-term problem,  and makes a series of improvements on the basis of KCF. LCT decomposes the task of tracking into translation and scale estimation of objects, adds re-detection framework and achieves a substantial increase in accuracy. For the long-term tracking task, we directly impose our latent constrains to LCT and generate a Latent Constrained Long-term Correlation Tracker (LC-LCT).

\begin{table}[h]
	\caption{Success and precision plots according to the online tracking benchmark \cite{benchmark} based on different features.}
	\begin{center}
	\begin{tabular}{m{2cm}<{\centering}|c|ccc}	
	\hline
	Feature& & Gray & HOG & VGG-19  \\
	\hline
	\multirow{2}[5]*{Precision Success}&KCF & 56.1\% & 74.0\% &89.1\%\\
	& LC-KCF &56.8\% &79.4\% &89.6\% \\
	\hline
	\multirow{2}[5]*{FPS}& KCF &   & 405.34 & 14.65(GPU) \\
	& LC-KCF &   & 337.41 & 14.04(GPU) \\
	\hline
	\end{tabular}
	\end{center}
	   			
	\label{table:feature}
\end{table}
	   		
\begin{table}[htbp]
	\centering
	\caption{Sensitivity experiments of parameter $T$, results surpass the baseline are marked in bold type.}
	\begin{tabular}{p{0.7cm}|p{0.48cm}<{\centering}p{0.48cm}<{\centering}p{0.48cm}<{\centering}p{0.48cm}<{\centering}p{0.48cm}<{\centering}p{0.48cm}<{\centering}p{0.48cm}<{\centering}p{0.48cm}<{\centering}}
	\hline
	T & 8    & 10    & 12    & 14    & 16    & 18    & 20    & 22   \\
	\hline
	\hline
    HOG   &  70.5\% & 72.7\% &\bf{76.4}\% &\bf{75.8}\%  &\bf{79.4\%}   & \bf{78.9\%} & \bf{77.5\%}  & \bf{76.7\%} \\
	\hline
	\end{tabular}%
	\label{table:parameter}%
\end{table}%
	
KCF and LCT  train two correlation filters (context tracker $R_{c}$ and target appearance tracker $R_{t}$) during the tracking process. Our latent constrains are only added to the context tracker (i.e., no change to the appearance tracker). Parameter settings follow \cite{longterm}. The results are shown in Fig. \ref{fig:longterm}
\begin{figure} [!t]
	\begin{center}
		\includegraphics[width=0.5\textwidth]{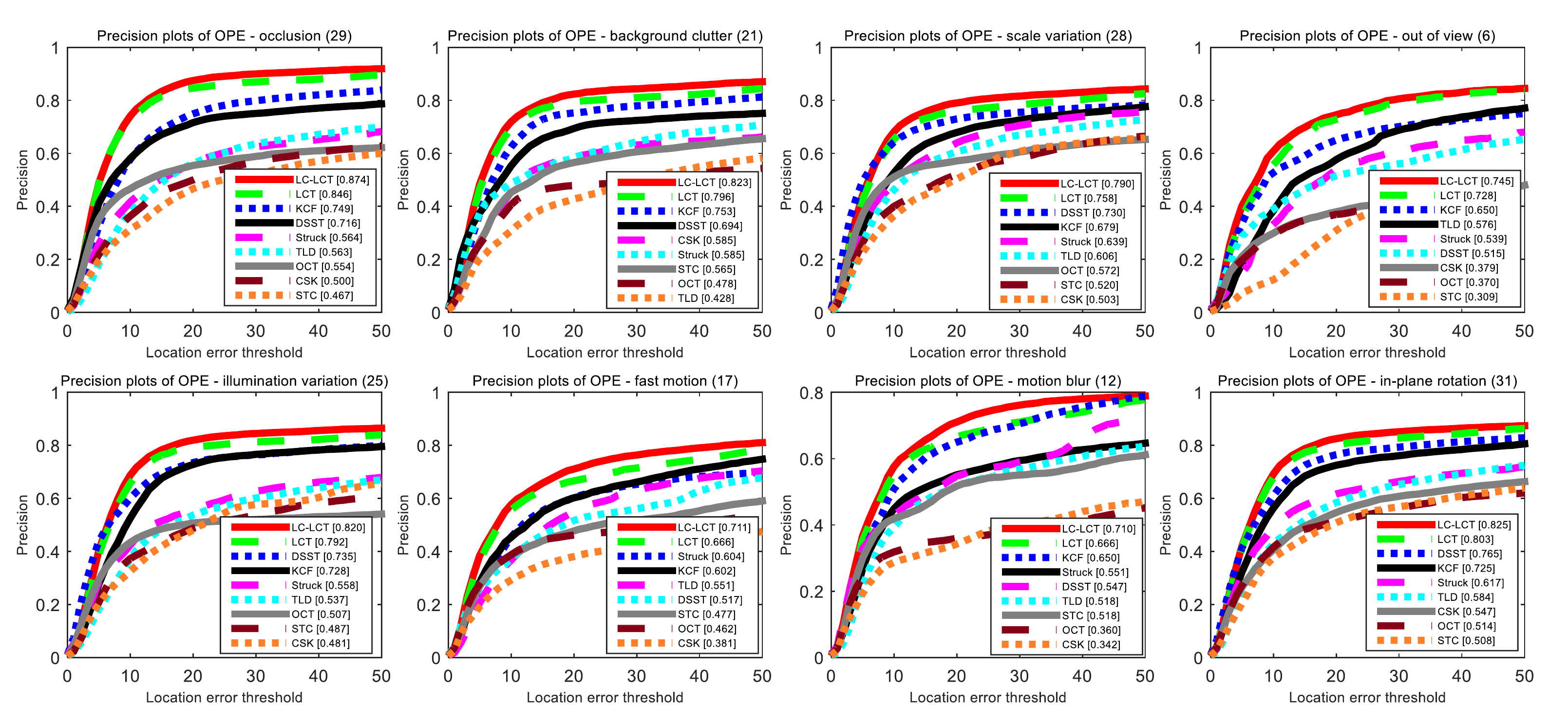}
	\end{center}
	\caption{Precision plots on some attribute categories.}
	\label{fig:occlusion}
\end{figure}
	
The LC-LCT and LCT achieve 78.2\% and 76.8\% based on the average success rate, while the KCF and DSST trackers respectively achieve 67.4\% and 67.8\%. In terms of precision, LC-LCT and LCT respectively achieve 86.5\% and 84.8\% when the threshold is set to 20. LC-LCT consistently obtains much higher tracking performance than KCF (74.0\%), DSST (73.7\%), Struck (65.6\%) and TLD (60.8\%).
{We also observe that LC-LCT exhibits very good performance on some attribute categories, such as occlusion, illumination variation, out of view, etc. On the subset of occlusion, LC-LCT is about $3$\% higher than LCT (87.4\% vs 84.6\%). These results are presented in Fig.\ref{fig:occlusion}.
In terms of tracking speed, LCT processes 29.67 frames per second (FPS), while LC-LCT has a processing rate of 25.43 FPS. Therefore, the proposed LC-LCT only has a minuscule frame rate drop as compared to the original LCT, yet is still able to achieve  real-time processing.
	
	

		
These results confirm that the latent constraint contributes to our tracker and enables it to perform better than the state-of-the-art methods.
	

\section{Conclusions}
In this paper, we have proposed the latent constrained  correlation filters (LCCF) and introduced a subspace ADMM algorithm to solve the new learning model. The theoretical analysis reveals that the new subspace ADMM is much faster than the original ADMM in terms of the convergence speed. The experimental results have shown consistent advantages over the state-of-the-arts when applying LCCF to several computer vision applications including eye detection, car detection and object tracking. {{In future work, we will incorporate the latent constraint in deep learning frameworks, and explore other applications such as action recognition~\cite{zhang2017action}, image retrieval~\cite{guo2017hash, lin2017hash} and visual saliency detection~\cite{zhang2017saliency, yao2017saliency}.}}


{	
\subsection*{Appendix I: Convergence of SADMM}

To prove the convergence of SADMM, we set
$F(\hat{\bf h}) = \frac{1}{2}\sum\limits_{i = 1}^B {||{{\hat{\bf y}}_i} - {{\hat{\bf X}}_i} \hat{\bf h}} ||_2^2 + \frac{\lambda }{2}||\hat{\bf h}||_2^2 $. Then Eq. 7 can be rewritten as:
\begin{equation}
	L_\sigma(\hat{\bf h})=F(\hat{\bf h})+ \frac{\sigma}{2} \lVert \hat{\bf h}-\hat{\bf g} \rVert^2.
	\label{eq:Appendix1}
\end{equation}
We set $\hat{\bf h}^*$ as the saddle point for the objective mentioned above. Considering the case $\varepsilon  = \| \hat{\mathbf{h}}^{t+1} - \hat{\mathbf{h}}^{t} \|^2$ as shown in Algorithm 1, $\hat{\bf h}^{t + 1}$ minimizes
\begin{equation}
	F(\hat{\bf h}^{t+1})+\frac{\sigma^{t}}{2} \lVert \hat{\bf h}^{k + 1}-\hat{\bf g}^{t} \rVert^2 + \frac{\sigma^{t}}{2} \| \hat{\mathbf{h}}^{t+1} - \hat{\mathbf{h}}^{t} \|^2.
	\label{eq:Appendix2}
\end{equation}
Since $\hat{\mathbf{g}}^{} = \hat{\mathbf{h}}^{}$, Eq.(\ref{eq:Appendix2}) is rewritten as:
\begin{equation}
	L_\sigma(\hat{\bf h})=F(\hat{\bf h}^{t+1})+ {\sigma^{t}} \lVert \hat{\bf h}^{t+1}-\hat{\bf g}^{t} \rVert^2,
	\label{eq:Appendix3}
\end{equation}
and the derivative of Eq. (\ref{eq:Appendix3}) is:
\begin{equation}
	\partial F(\hat{\bf h}^{t+1})+2\sigma^{t}( \hat{\bf h}^{t+1}-\hat{\bf g}^{t} ),
	\label{eq:Appendix4}
\end{equation}
which can also be considered as the derivative of Eq.(\ref{eq:Appendix5}):
\begin{equation}
	F(\hat{\bf h})+2\sigma^{t}( \hat{\bf h}^{t+1}-\hat{\bf g}^{t} ) \hat{\bf h}.
	\label{eq:Appendix5}
\end{equation}
We have:
\begin{equation}
    \begin{aligned}
	&F(\hat{\bf h}^{t+1})+2\sigma^{t}( \hat{\bf h}^{t+1}-\hat{\bf g}^{t} )\hat{\bf h}^{t+1} \\
    &\le  F(\hat{\bf h}^*)+2\sigma^{t}( \hat{\bf h}^{t+1}-\hat{\bf g}^{t} ) \hat{\bf h}^*
	\label{eq:Appendix6}
    \end{aligned}
\end{equation}
and obtain:
\begin{equation}
	F(\hat{\bf h}^{t+1})-F(\hat{\bf h}^*)\le 2\sigma^{t}(\hat{\bf h}^{t+1}-\hat{\bf g}^{t})(\hat{\bf h}^{*}-\hat{\bf h}^{t+1}).
	\label{eq:Appendix7}
\end{equation}
In addition, based on Eq.(\ref{eq:Appendix1}), we have:
\begin{equation}
	F(\hat{\bf h}^{t+1})+ \frac{\sigma^{t}}{2}  \lVert \hat{\bf h}^{t+1} -\hat{\bf g}^{t} \rVert^2 \ge F(\hat{\bf h}^{*})+ \frac{\sigma^{t}}{2} \sigma^{t} \lVert \hat{\bf h}^{*} -\hat{\bf g}^{t}\rVert^2.
	\label{eq:Appendix8}
\end{equation}
Since $\hat{h}^*$ is the saddle point for our problem,
%
we have:
\begin{equation}
	F(\hat{\bf h}^{*})-F(\hat{\bf h}^{t+1}) \le \frac{\sigma^{t}}{2} *( \lVert \hat{\bf h}^{t+1} -\hat{\bf g}^{t} \rVert^2 - \lVert \hat{\bf h}^{*} -\hat{\bf g}^{t}\rVert^2).
	\label{eq:Appendix9}
\end{equation}
From Eq. (\ref{eq:Appendix7}) and Eq. (\ref{eq:Appendix9}), we get:
\begin{equation}
	\begin{aligned}
	&-4 \lVert \hat{\bf h}^{t+1} - \hat{\bf h}^* \rVert^2 + 4(\hat{\bf h}^*-\hat{\bf g}^{t})(\hat{\bf h}^*-\hat{\bf h}^{t+1}) \\
		&+\lVert \hat{\bf h}^{t+1} - \hat{\bf g}^{t} \rVert^2 - \lVert \hat{\bf h}^* - \hat{\bf g}^{t} \rVert^2 \ge 0.
		\label{eq:Appendix10}
	\end{aligned}
\end{equation}
where we also used
$4(\hat{\bf h}^{t+1} - \hat{\bf h}^* + \hat{\bf h}^* - \hat{\bf g}^{t})(\hat{\bf h}^* - \hat{\bf h}^{t+1})$ to change the right part of  Eq.  (\ref{eq:Appendix7}). And then we have:

\begin{equation}
\begin{aligned}
  \lVert \hat{\bf h}^{t+1} - \hat{\bf g}^{t} \rVert^2 = &\lVert \hat{\bf h}^{t+1} - \hat{\bf h}^*\rVert^2 + \rVert \hat{\bf h}^* - \hat{\bf g}^{t} \rVert^2 \\
  & + 2 (\hat{\bf h}^{t+1} - \hat{\bf h}^*)(\hat{\bf h}^* - \hat{\bf g}^{t}),
  \end{aligned}
  \label{eq:Appendix11}
\end{equation}
  we have:
\begin{equation}
	\begin{aligned}
		&-4\lVert \hat{\bf h}^{t+1} - \hat{\bf h}^* \rVert^2 + 4(\hat{\bf h}^* - \hat{\bf g}^{t})(\hat{\bf h}^* - \hat{\bf h}^{k+1}) \\&+ \lVert \hat{\bf h}^{t+1} - \hat{\bf g}^{t} \rVert^2
		+\lVert \hat{\bf h}^{t+1} - \hat{\bf g}^{t} \rVert^2 \\
		&- \lVert \hat{\bf h}^* - \hat{\bf g}^{t} \rVert^2- \lVert \hat{\bf h}^{t+1} - \hat{\bf g}^{t} \rVert^2 \\
		=&-4 \lVert \hat{\bf h}^{t+1} - \hat{\bf h}^* \rVert^2 + 2\lVert \hat{\bf h}^{t+1} - \hat{\bf h}^* \rVert^2 + 2\lVert \hat{\bf h}^* - \hat{\bf g}^{t} \rVert^2\\
		-& \lVert \hat{\bf h}^* - \hat{\bf g}^{t} \rVert^2 - \lVert \hat{\bf h}^{t+1} - \hat{\bf g}^{t} \rVert^2 \ge 0.
		\label{eq:Appendix12}
	\end{aligned}
\end{equation}
Therefore we obtain:
\begin{equation}
	-2\lVert \hat{\bf h}^{t+1} - \hat{\bf h}^* \rVert^2 + \lVert \hat{\bf h}^* - \hat{\bf g}^{t} \rVert^2 - \lVert \hat{\bf h}^{t+1} - \hat{\bf g}^{t} \rVert^ \ge 0
	\label{eq:Appendix13}
\end{equation}
Again, using $\hat{\mathbf{g}}^{} = \hat{\mathbf{h}}^{}$, we have:
\begin{equation}
	\lVert \hat{\bf h}^{t+1} - \hat{\bf h}^* \rVert^2 \le  \frac{1}{2} \lVert \hat{\bf h}^* - \hat{\bf h}^{t} \rVert^2 -  \frac{1}{2} \lVert \hat{\bf h}^{t+1} - \hat{\bf g}^{t} \rVert^2.
	\label{eq:Appendix14}
\end{equation}
Compared to \cite{admm}, our method is as efficient as ADMM in terms of convergence speed.

\subsection*{Appendix II: Kernelized correlation filters (KCF)}
	
We briefly review the KCF algorithm. {{Based on the input image of $\mathbf x$ of $M \times N$ pixels,  in the spatial domain, KCF is described as}}
\begin{equation}
	\begin{aligned}
	&\underset{{\bf h}}{\min}
	& &\sum_{i}{{\xi}}_i^2 \\
	&\textit{subject to}
	& &{{y}}_i-{{\bf h}}^T\phi({{\bf x}}_i)={{\xi}}_i     ~~\forall{i};
	& & ||{\bf h}||\le C.&
	\end{aligned}
	\label{eq:KCF1}
\end{equation}
	
{{In Eq. (\ref{eq:KCF1}), ${\bf x}_i$ is the  circular samples of image ${\bf x}$. ${\bf y}$ representing a Gaussian output and ${\bm\xi}$ denoting a slack variable, are vectors. Therefore,  ${y}_i$ and ${\xi}_i$ in Eq. (\ref{eq:KCF1}) are scalars.}}
$\phi(.)$ (later ${\bm\phi}_i$) refers to the nonlinear kernel function,
and $C$ is a small constant. Then, we have
\begin{equation}
	\begin{aligned}
		&\mathcal{L}_p=\sum_{i=1}^{M\times N}{\xi}_i^2+\sum_{i=1}^{M\times N}{\theta}_i({y}_i-{{\bf h}}^T{{\bm\phi}}_i-{\xi}_i)+\lambda(||{\bf h} ||^2-C^2),
	\label{eq:KCF2}
	\end{aligned}
\end{equation}
and KKT conditions are obtained:
\begin{equation}
	{\theta}_i=2{\xi}_i
	\label{eq:KCF3}
\end{equation}
\begin{equation}
	{{\bf h}}=\sum_i^{M\times N}{\frac{{\theta}_i}{2\lambda}}{\bm\phi_i}.
	\label{eq:KCF4}
\end{equation}
Based on Eq. (\ref{eq:KCF4}), by setting ${\alpha}_i=\frac{{\theta}_i}{2\lambda}$,  solving ${\bf h}$ in the frequency domain will be converted to solving a dual variable $\bm\alpha$ in a dual space. Plugging $\bm\alpha$ back into Eq.\eqref{eq:KCF2}, we formulate a new maximizing objective function $\textit{E}(\bm\alpha)$:	
\begin{equation}
	\textit{E}(\bm\alpha)=-\lambda^2\sum_{i=1}^{M\times N}{{\alpha}_i^2}+2\lambda\sum_{i=1}^{M\times N}{{\alpha}_i{y}_i}-\lambda\sum_{i,j=1}^{M\times N}{{\alpha}_i{\alpha}}_j{\bf K}_{i,j}-\lambda C^2,
	\label{eq:KCF5}
\end{equation}	
where $\bf K$ is a kernel matrix. {{Then, the solution of Eq. (\ref{eq:KCF5})}} is
\begin{equation}
	{\bm\alpha}=({{\bf K}+\lambda \bf I})^{-1}{\bf y}.
	\label{eq:KCF6}
\end{equation}
According to the properties of the cyclic matrix, Eq. (\ref{eq:KCF6}) is transformed into the Fourier domain to speed up the calculation based on  Fast Fourier Transform (FFT). Then, we have $\hat{\bf \alpha}$ as:	
\begin{equation}
	\hat{\bm\alpha}=({{\bf K}^{xx}+\lambda \bf I})^{-1}{\hat{\bf y}},
	\label{eq:KCF7}
\end{equation}	
where ${\bf K}^{xx}$ refers to the first row of ${\bf K}$.

\begin{IEEEbiographynophoto}{Baochang Zhang} is currently an associate professor with Beihang University, Beijing, China. His research interests include pattern recognition, machine learning, face recognition, and wavelets.
\end{IEEEbiographynophoto}
\begin{IEEEbiographynophoto}{Shangzhen Luan} received the B.S. degrees in automation from Beihang University. His research interests include signal and image processing, pattern recognition and computer vision
\end{IEEEbiographynophoto}
\begin{IEEEbiographynophoto}{Chen Chen} is currently a Postdoctoral Fellow with the Center for Research in Computer Vision, University of Central Florida, Orlando, FL, USA. His research interests include compressed sensing, signal and image processing, pattern recognition, and computer vision.
\end{IEEEbiographynophoto}
%
\begin{IEEEbiographynophoto}{Jungong Han} is currently a tenured faculty member with the School of Computing and Communications at Lancaster University, UK. His research interestes include video analysis, computer vision and artificial intelligence.
\end{IEEEbiographynophoto}
%

\begin{IEEEbiographynophoto} {Wei Wang} is currently an Associate Professor with the School of Automation Science and Electrical Engineering at Beihang University and supported by BUAA Young Talent Recruitment Program.
\end{IEEEbiographynophoto}

\begin{IEEEbiographynophoto}{Alessandro Perina} is currently a data scientist at Microsoft Corporation, Redmond, WA, USA. His area of expertise lies in machine learning, statistics and applications, particularly in probabilistic graphical models, unsupervised learning, and optimization algorithms.
\end{IEEEbiographynophoto}
%
\begin{IEEEbiographynophoto}{Ling Shao} (M’09–SM’10) is currently a Professor with the School of Computing Sciences, University of East Anglia, Norwich, U.K. His research interests include computer vision, image/video processing, and machine learning. He is a fellow of the British Computer Society and the Institution of Engineering and Technology.
\end{IEEEbiographynophoto}

\end{document}